\documentclass[conference]{IEEEtran}
\IEEEoverridecommandlockouts
\usepackage{cite}
\usepackage{amsmath,amssymb,amsfonts}
\usepackage{algorithmic}
\usepackage{graphicx}
\usepackage{textcomp}
\usepackage{xcolor}

\usepackage{booktabs,multirow,subfig,hyperref}
\usepackage[marginal]{footmisc}

\def\BibTeX{{\rm B\kern-.05em{\sc i\kern-.025em b}\kern-.08em
    T\kern-.1667em\lower.7ex\hbox{E}\kern-.125emX}}
\begin{document}

\title{DDLNet: Boosting Remote Sensing Change Detection with Dual-Domain Learning\\
\thanks{This work was supported by the Public Welfare Science and Technology Plan of Ningbo City (2022S125), the Key Research and Development Plan of Zhejiang Province (2021C01031) and the Science and Technology Innovation 2025 Major Project of Ningbo City (Grant No. 2022Z032). }
}

\author{\IEEEauthorblockN{1\textsuperscript{st} Xiaowen Ma$^{\dag}$}
\IEEEauthorblockA{\textit{School of Software Technology} \\
\textit{Zhejiang University}\\
Hangzhou, China \\
xwma@zju.edu.cn}
\and
\IEEEauthorblockN{2\textsuperscript{nd} Jiawei Yang$^{\dag}$}
\IEEEauthorblockA{\textit{School of Software Technology} \\
\textit{Zhejiang University}\\
Hangzhou, China \\
jw.yang@zju.edu.cn}
\and
\IEEEauthorblockN{3\textsuperscript{rd} Rui Che}
\IEEEauthorblockA{\textit{School of Software Technology} \\
\textit{Zhejiang University}\\
Hangzhou, China \\
rche@zju.edu.cn}
\and
\IEEEauthorblockN{4\textsuperscript{th} Huanting Zhang}
\IEEEauthorblockA{\textit{School of Software Technology} \\
\textit{Zhejiang University}\\
Hangzhou, China \\
ht.zhang@zju.edu.cn}
\and
\IEEEauthorblockN{5\textsuperscript{th} Wei Zhang$^{*}$}
\IEEEauthorblockA{\textit{School of Software Technology} \\
\textit{Zhejiang University}\\
Hangzhou, China \\
\textit{Innovation Center of Yangtze River Delta}
\textit{Zhejiang University}\\
Jiaxing Zhejiang, 314103, China \\
cstzhangwei@zju.edu.cn}
\thanks{\dag \ Equal contributions. * Corresponding author.}
}

\maketitle

\begin{abstract}
Remote sensing change detection (RSCD) aims to identify the changes of interest in a region by analyzing multi-temporal remote sensing images, and has an outstanding value for local development monitoring. Existing RSCD methods are devoted to contextual modeling in the spatial domain to enhance the changes of interest. Despite the satisfactory performance achieved, the lack of knowledge in the frequency domain limits the further improvement of model performance. In this paper, we propose DDLNet, a RSCD network based on dual-domain learning (i.e., frequency and spatial domains). In particular, we design a Frequency-domain Enhancement Module (FEM) to capture frequency components from the input bi-temporal images using Discrete Cosine Transform (DCT) and thus enhance the changes of interest. Besides, we devise a Spatial-domain Recovery Module (SRM) to fuse spatiotemporal features for reconstructing spatial details of change representations. Extensive experiments on three benchmark RSCD datasets demonstrate that the proposed method achieves state-of-the-art performance and reaches a more satisfactory accuracy-efficiency trade-off. Our code is publicly available at~\href{https://github.com/xwmaxwma/rschange}{https://github.com/xwmaxwma/rschange}.
\end{abstract}


\section{Introduction}
RSCD is a crucial task to identify the change of interest, represented by spectral variations, by analyzing multi-temporal remote sensing images captured over the same geographic region. As a result, image pixels are classified into a change map to determine whether each part of the region involves the change of interest. This analytical process reveals the local development patterns from both natural and socio-economic perspectives, and possesses significant implications for urban planning, environmental monitoring, and disaster assessment~\cite{feng2023compact}.

The rapid advancement of deep learning techniques has significantly propelled the area of RSCD, particularly with the progress of Convolutional Neural Networks (CNNs) in extracting hierarchical features~\cite{xu2015neural}. For instance, LGPNet~\cite{liu2021building} incorporates the attention mechanism to enhance the expressiveness of change representations. The Siamese-based STANet~\cite{chen2020spatial} adopts the BAM to leverage global spatio-temporal relations for discriminative features and the PAM to aggregate multi-scale attentional representations for the details on fine-grained objects. DASNets~\cite{chen2020dasnet} employs a dual-attention mechanism to capture long-range dependencies and obtain discriminative feature representations for change detection. SNUNet~\cite{fang2021snunet} integrates a Siamese neural network with the U-Net architecture to preserve high-level semantic information and detailed features through jump connections. ChangeFormer~\cite{bandara2022transformer} introduces a Transformer-based Siamese neural network, which exploits the Transformer model for spatial relationship learning in temporal sequences. Reaching the considerable improvements in the performance, these methods primarily focus on solving the problem of RSCD only in the spatial domain, which leaves room for further advancements.

More recently, extensive attempts from the perspective of the frequency domain have been undertaken in the areas of target detection and semantic segmentation. For example, incorporating the features in the frequency domain features after transforming images into CNNs has been validated to improve the performance of image analysis tasks~\cite{xu2020learning}. AFFormer~\cite{dong2023afformer} learns local descriptive representations of clustering prototypes from a frequency perspective, eliminating the need for a complex decoder. Simultaneously, it incorporates a Transformer with linear complexity, simplifying semantic segmentation to conventional classification. These studies have shown that the introduction of frequency information contributes to the extraction of discriminative features for semantic recognition. Therefore, we are inspired to explore from the frequency domain to improve the performance of RSCD.

\begin{figure*}[htb]
  \centering
  \includegraphics[width=0.9\textwidth]{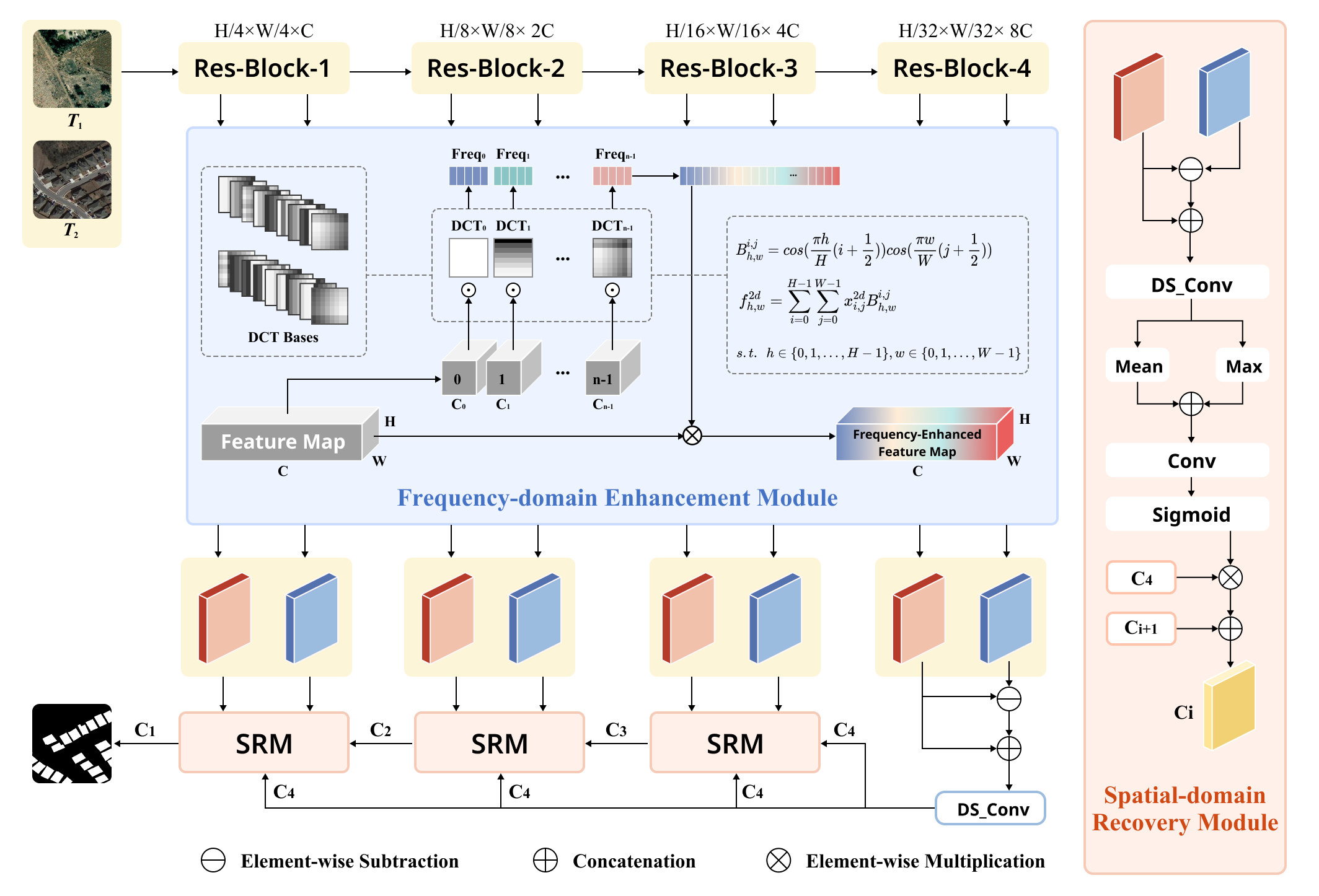}
  \caption{Architecture of the proposed DDLNet, which consists of four components: a Siamese backbone, FEM, SRM and a lightweight change detection header. DS\_Conv refers to the depth-wise separable convolution, which is employed to reduce the number of parameters and computational cost while minimizing the compromise in the performance.}
  \label{Fig.1}
\end{figure*}

In this paper, we propose DDLNet, a RSCD network based on dual-domain learning, which involves Frequency-domain Enhancement Modules (FEM) and Spatial-domain Recovery Modules (SRM). Given the feature maps extracted by a Siamese neural network from the input bi-temporal images, the FEM captures the frequency components using the Discrete Cosine Transform (DCT) to enhance the changes of interest. Meanwhile, SRM at different scales facilitate the recovery of spatial details through the fusion of coarse and refined representations on the change of interest in a cascaded manner. In addition, we employ a lightweight decoder to interconnect the representations across all scales to generate the change map.

The primary contributions of this paper can be summarized as follows:
\begin{itemize}
  \item we introduce frequency-domain enhancement that extracts frequency components from bi-temporal image feature maps and enhances the changes of interest,
  \item we enable spatial-domain recovery that employs cross-time and cross-scale mechanisms for feature fusion and reconstructs spatial details in change representations,
  \item we propose a novel RSCD method that achieves state-of-the-art performance on three change detection datasets while reaching a more satisfactory trade-off between accuracy and efficiency.
\end{itemize}

\section{Method}

Fig.\ref{Fig.1} illustrates the architecture of our DDLNet that consists of four key components on feature extraction, frequency-domain enhancement, spatial-domain recovery, and decoding. Given the input bi-temporal images $T_1$ and $T_2$, a Siamese neural network (i.e, two-stream ResNet-18) is utilized to extract bi-temporal feature maps at four different scales. At each scale, the bi-temporal feature maps from the corresponding residual block are enhanced regarding the changes of interest between $T_1$ and $T_2$ in the frequency domain using a FEM, which leverages the DCT to effectively capture frequency components. The frequency-enhanced feature maps at the fourth scale experiences an initial feature fusion, followed by being processed by the SRM, which focuses on recovering the spatial details on the change of interest, along with the feature maps at each other scale. The feature maps at all scales are upsampled and concatenated along the channel direction; the number of channels count are reinstated through convolution. Finally, an upsampling yields a map on the change of interest of the same size as $T_1$ and $T_2$. 

\subsection{Frequency-domain Enhancement Module}

Conventional RSCD methods concentrate on exploiting the spatial domain and leave a significant space to explore the potential of the frequency domain. A wide spectrum of previous studies have demonstrated the effectiveness of considering the frequency domain in extracting discriminative features and achieving accurate semantic recognition. Inspired by AFFormer~\cite{dong2023afformer} and Fcanet~\cite{qin2021fcanet}, we design the FEM that adopts the DCT to efficiently capture the frequency components within features and thus enhance the distinguishness of the changes of interest in the input images.

The two-dimensional DCT can be formulated as follows:

\begin{equation}
B_{h,w}^{i,j}=cos(\frac{\pi h}H(i+\frac12))\cos(\frac{\pi w}W(j+\frac12)),
\end{equation}

\begin{equation}
f_{h,w}^{2d}=\sum_{i=0}^{H-1}\sum_{j=0}^{W-1}x_{i,j}^{2d}B_{h,w}^{i,j},
\label{equation:1}
\end{equation}
where $f^{2d} \in \mathbb{R}^{H \times W}$ denotes the 2D DCT frequency spectrum, $x^{2d} \in \mathbb{R}^{H \times W}$ represents the input, $H$ and $W$ refer to the height and the width of $x^{2d}$, respectively. Additionally, in this formula, $h\in\{0,1,\cdots,\textit{H}-1\}$, $w\in\{0,1,\cdots,\textit{W}-1\}$, $i\in\{0,1,\cdots,\textit{H}-1\}$, and $j\in\{0,1,\cdots,\textit{W}-1\}$.

Our FEM divides the input image $X$ along the channel dimension into parts $[X^0,X^1,\cdots,X^{n-1}]$, where $X^i\in\mathbb{R}^{C'\times H\times W},i\in\{0,1,\cdots,n-1\}$ and $C'=\frac{C}{n}$ under the assumption that $C$ is divisible by $n$. The corresponding 2D DCT frequency component is assigned to each part $X^{i}$, so as to obtain the frequency vector via concatenation as follows:
\begin{equation}
Freq= \operatorname{cat}([Freq^{0},Freq^{1},\cdots,Freq^{n-1}]), 
\label{equation:2}
\end{equation}
where each frequency component is formulated as follows:
\begin{equation}
Freq^{i} =\mathrm{DCT}^{u_{i},v_{i}}(X^{i}) =\sum_{h=0}^{H-1}\sum_{w=0}^{W-1}X_{h,w}^{i}B_{h,w}^{u_{i},v_{i}},
\end{equation}
where $\textit{Freq}^i \in \mathbb{R}^{C'}$ is a $C'$-dimensional vector, $[u_i,v_i]$ denote the frequency component's 2D indices for $X^i$, and $i\in\{0,1,\cdots,\textit{H}-1\}$. The FEM's output is then calculated as follows:

\begin{equation}
\mathcal{Z}=Sigmoid(fc(Freq)).
\label{equation:3}
\end{equation}

in which $fc$ represents the one-dimensional convolution, and $\mathcal{Z}$ denotes the frequency-enhanced feature map. As described by Equation~\ref{equation:2} and~\ref{equation:3}, multiple frequency components are obtained so as to expedite the effective enrichment of the representation on the changes of interest for feature enhancement.

Furthermore, we need to select frequency component indices [$u_{i}$, $v_{i}$] for each part $X^{i}$. Specifically, we first determine the importance of each frequency component and then investigate the impact of using different numbers of frequency components. Following the previous work~\cite{qin2021fcanet}, we conduct pre-training on ImageNet to ascertain the significance of different frequency components for classification accuracy. We selected the top-n values as the target frequencies. The specific effects of different numbers of frequency components will be discussed in the experimental section. In addition, it is noteworthy we omit the normalization factor for simplicity, as shown in Equation~\ref{equation:1}, which however barely affects the performance of the proposed method.

\subsection{Spatial-domain Recovery Module}

Our SRM achieves feature fusion through cross-time and cross-scale mechanisms. Specifically, following the frequency enhancement of features at each hierarchical level, the deepest-layer features undergo element-wise subtraction and are concatenated with the image captured at time T1, resulting in a representation of deep-level changes. Subsequently, the representation of deep-level changes is independently fused with shallow-level features across scales. This enables the utilization of higher-level change representations to guide context modeling for lower-level representations, enhancing semantic information in the lower-level features and improving the recovery of spatial details for the change representations.

Given the frequency-enhanced feature maps $\mathcal{Z}_1$ and $\mathcal{Z}_2$ output by the FEM from $T_1$ and $T_2$ at each of the first three scales, SRM first conducts an initial feature fusion to reach a coarse representation $\mathcal{Z}_c$ on the change of interest as follows:

\begin{equation}
\mathcal{Z}_c =\psi ((\mathcal{Z}_1 \ominus \mathcal{Z}_2) \oplus \mathcal{Z}_1),
\end{equation}
where $\ominus$ denotes element-wise subtraction, and $\psi$ refers to depth-wise separable convolution. The weight map $\mathcal{W}_c$ of $\mathcal{Z}_c$ are calculated as follows:

\begin{equation}
\mathcal{W}_1 =Sigmoid (\phi (\mathcal{P}_{mean}(\mathcal{Z}_c) \oplus \mathcal{P}_{max}(\mathcal{Z}_c)),
\end{equation}
where $\mathcal{P}_{mean}$ and $\mathcal{P}_{max}$ respectively denote average pooling and max pooling, and $\phi$ represents $1\times 1$ convolution. The weight map is then applied to the coarse representation $\mathcal{C}_{4}$ at the fourth scale, which reaches the refined representation on the change of interest $\mathcal{C}_i$ as follows:

\begin{equation}
\mathcal{C}_i =(\mathcal{W}_1 \otimes \mathcal{C}_{4}) \oplus \mathcal{C}_{i + 1},
\end{equation}
where $\mathcal{C}_{i + 1}$ represents the refined representation on the change of interest from the previous SRM.

\begin{table*}[htb]
 \centering
 \caption{
      Comparison of our DDLNet and representative RSCD methods regarding performance on three datasets. Highest scores are in bold. All scores are reported in percentage.
 }
  \resizebox{\textwidth}{26mm}{
 \label{tab:1}
 \begin{tabular}{l|ccccc|ccccc|ccccc}
  \toprule
  \multirow{2}{*}{\textbf{Method}} &\multicolumn{5}{c|}{\textbf{WHU-CD}} &\multicolumn{5}{c|}{\textbf{LEVIR-CD}} &\multicolumn{5}{c}{\textbf{CLCD}}\\
     &F1 &Pre. &Rec. &IoU &OA  &F1 &Pre. &Rec. &IoU &OA &F1 &Pre. &Rec. &IoU &OA \\
  \midrule
  FC-EF~\cite{daudt2018fully}  &72.01 &77.69 &67.10 &56.26 &92.07 &83.40 &86.91 &80.17 &71.53 &98.39 &48.64 &73.34 &36.29 &32.14 &94.30\\
  FC-Siam-Di~\cite{daudt2018fully}  &58.81 &47.33 &77.66 &41.66 &95.63 &86.31 &89.53 &83.31 &75.92 &98.67 &44.10 &72.97 &31.60 &28.29 &94.04\\
  FC-Siam-Conc~\cite{daudt2018fully} &66.63 &60.88 &73.58 &49.95 &97.04 &83.69 &91.99 &76.77&71.96&98.49&54.48&68.21&45.22&37.35&94.35\\
  IFNet~\cite{zhang2020deeply} &83.40 &\bf96.91 &73.19 &71.52 &98.83 &88.13 &\bf94.02 &82.93&78.77&98.87&48.65&49.96&47.41&32.14&92.55\\
  DTCDSCN~\cite{liu2020building}&71.95 &63.92 &82.30 &56.19 &97.42 &87.67 &88.53&86.83&78.05&98.77&60.13&62.98&57.53&42.99&94.32 \\
  BIT~\cite{chen2021remote}&83.98 &86.64 &81.48 &72.39 &98.75 &89.31 &89.24 &89.37&80.68&98.92&57.13&64.39&51.34&39.99&94.27\\
  LGPNet~\cite{liu2021building}&79.75 &89.68 &71.81 &66.33 &98.33 &89.37 &93.07 &85.95&80.78&99.00&55.42&62.98&49.49&38.33&94.10 \\
  SNUNet~\cite{fang2021snunet}&83.50 &85.60 &81.49 &71.67 &98.71 &88.16 &89.18 &87.17&78.83&98.82&60.54&65.63&56.19&43.41&94.55\\
  DMATNet~\cite{song2022remote}&85.07 &89.46 &82.24 &74.98 &95.83 &89.97 &90.78 &89.17&81.83&98.06&66.56&72.74&61.34&49.87&95.41\\
  ChangeStar(FarSeg)~\cite{zheng2021change}&87.01 &88.78 &85.31 &77.00 &98.70 &89.30 &89.88 &88.72 &80.66 &98.90 &60.75 &62.23 &59.34 &43.63 &94.30 \\
  ChangeFormer~\cite{bandara2022transformer}&81.82 &87.25 &77.03 &69.24 &94.80 &90.40 &92.05 &88.80 &82.48 &99.04 &58.44 &65.00 &53.07 &41.28 &94.38 \\
  USSFC-Net~\cite{lei2023ultralightweight}&88.40 &90.67 &86.24 &79.21 &98.96 &90.11 &88.88 &89.35&80.37&98.89&63.04&64.83&61.34&46.03&94.42\\
  \midrule
  DDLNet (Ours)&\bf90.56 &91.56 &\bf90.03 &\bf82.75 &\bf99.13 &\bf90.60 &91.72 &\bf89.41 &\bf82.49 &\bf99.10&\bf72.44&\bf78.93&\bf66.90&\bf56.78&\bf95.94\\
  \bottomrule
 \end{tabular}}
\end{table*}

\subsection{Loss Function}

Considering the renowned imbalance problem on the distribution of changed and unchanged pixels in RSCD, we devise a hybrid loss function integrating both focal loss and dice loss, whose formulation is as follows:
\begin{equation}
\mathcal{L}=\mathcal{L}_{focal}+\mathcal{L}_{dice}.
\end{equation}
Specifically, the focal loss is formulated as follows:
\begin{equation}
\mathcal{L}_{\mathrm{focal}}=-\alpha(1-\hat{p})^{\gamma}log(\hat{p}),
\end{equation}

\begin{equation}
\left.\hat{p}=\left\{\begin{array}{ll}p,&\text{if}\quad y=1\\1-p,&\text{otherwise}\end{array}\right.\right. ,
\end{equation}
where $\alpha $ and $\gamma$ represent two hyperparameters governing the contribution of positive and negative samples and the method's attention to the challenging samples, respectively (We set $\alpha $ to 0.2 and $\gamma$ to 2 in our experiments), $p$ denotes the probability, and $y$ refers to the binary label for unchanged pixels (0) and changed pixels (1). Besides, the dice loss is formulated as follows:

\begin{table}[ht]
 \centering
 \caption{
      Comparison of our DDLNet and its variants regarding RSCD performance on the WHU dataset. Highest scores are in bold. All scores are reported in percentage.
 }
  % \resizebox{\textwidth}{12mm}{
 \label{tab:2}
 \setlength{\tabcolsep}{3mm}{
 \begin{tabular}{l|ccccc}
  \toprule
  \multirow{0.9}{*}{\textbf{Model}} &F1 &Pre. &Rec. &IoU &OA \\
  \midrule
  Base  &87.96 &88.01 &88.03 &78.73 &98.69 \\
  Base+FEM  &89.17 &89.96 &88.41 &80.46 &98.98 \\
  Base+SRM &89.38 &88.75 &90.01 &80.79 &98.97 \\
  \midrule
  DDLNet (Ours)&\bf90.56 &\bf91.56 &\bf90.03 &\bf82.75 &\bf99.13 \\
  \bottomrule
 \end{tabular}}
\end{table}

\begin{equation}
\mathcal{L}_{dice}=1-\frac{2\cdot E\cdot Softmax(E^{\prime})}{E+Softmax(E^{\prime})},
\end{equation}

\begin{equation}
E^{\prime}=\left\{e_k^{\prime}\in\mathbb{R}^2|k=1,2,\ldots,H\times W\right\},
\end{equation}
where $E$ denotes the ground truth, and $E^{\prime}\in\mathbb{R}^{H\times W\times2}$ represents the change map including labeled pixels $\{e_k^{\prime}\}$.

\section{Experimental}
\subsection{Datasets and Evaluation Metrics}
We compare the proposed DDLNet on the three RSCD datasets by five commonly used metrics: F1 score (F1), recall (Rec), precision (Pre), intersection over union (IoU), and overall accuracy (OA).

The WHU architectural dataset~\cite{ji2018fully} is a public CD dataset containing a pair of bit-time aerial images of 32507×15354 dimensions. Following previous work~\cite{chen2021remote}, we crop the images into patches of 256×256 size and randomly divide them into a training set (6096 images), a validation set (762 images), and a test set (762 images).

The LEVIR-CD dataset~\cite{chen2020spatial} is a CD dataset containing 637 pairs of high-resolution dual-temporal remote sensing images. we crop the images into non-overlapping patches of 256 × 256 size, and we randomly divide them into a training set (7120 images), a validation set (1024 images), and a test set (2048 images).

\begin{table}[ht]
 \centering
 \caption{
      The performance of DDLNet on the WHU-CD and LEVIR-CD datasets when using different frequency components. $n$ represents the number of frequency components.
 }
 \setlength{\tabcolsep}{5mm}{
 \label{tab:3}
 \begin{tabular}{l|cc|cc}
  \toprule
  \multirow{2}{*}{\textbf{n}} &\multicolumn{2}{c|}{\textbf{WHU-CD}} &\multicolumn{2}{c}{\textbf{LEVIR-CD}} \\
     &F1 &IoU &F1 &IoU \\
  \midrule
  4 &90.20 &82.16 &90.30 &82.31\\
  8 &88.03 &78.62 &90.27 &82.27\\
  16 &\bf90.56 &\bf82.75 &\bf90.60 &\bf82.46\\
  32 &89.11 &80.35 &90.15 &82.07\\
  \bottomrule
 \end{tabular}}
\end{table}

The CLCD dataset~\cite{liu2022cnn} consists of 600 pairs of sample images of farmland changes with a size of 512 × 512 and it is randomly divided into a training set (360 images), a validation set (120 images) and a test set (120 images).

\subsection{Implementation Details}

We implement the proposed DDLNet using Python and Pytorch on a workstation equipped with two NVIDIA GTX A5000 graphics cards (48 GB of GPU memory in total.) We set the base learning rate to 0.05 and optimize the model using stochastic gradient descent (SGD) with a momentum of 0.9. and weight decay to 5e-5. Also, during training, we update the learning rate using a multi-step learning rate decay strategy with a gamma value of 0.1. In the experiments, the WHU-CD dataset and LEVIR-CD dataset batch sizes are set to 8, and the CLCD dataset batch size is set to 4. We perform data augmentation, such as rotation and flipping, on the input training data.

\begin{table}[ht]
 \centering
 \caption{
      Comparison of our DDLNet and representative RSCD methods regarding the number of parameters (Params) in million (M) and the number of Flops in Giga (G). Highest scores are in bold. The second-highest scores are underscored. All scores are reported in percentage.
 }
  % \resizebox{\textwidth}{12mm}{
 \label{tab:4}
 \setlength{\tabcolsep}{7mm}{
 \begin{tabular}{l|ccccc}
  \toprule
  \multirow{1}{*}{\textbf{Model}} &\textbf{Params (M)} &\textbf{Flops (G)} \\
  \midrule
  IFNet~\cite{zhang2020deeply}  &50.44 &82.26 \\
  DTCDSCN~\cite{liu2020building}  &41.07 &\underline{14.42}\\
  LGPNet~\cite{liu2021building} &70.99 &125.79 \\
  SNUNet~\cite{fang2021snunet} &\bf12.03 &54.88 \\
  DMATNet~\cite{song2022remote} &13.27 &- \\
  ChangeFormer~\cite{bandara2022transformer} &41.03 &202.79 \\
  \midrule
  DDLNet (Ours) &\underline{12.67} &\bf7.35 \\
  \bottomrule
 \end{tabular}}
\end{table}

\begin{figure*}[htb]
	\centering
	\begin{minipage}{0.11\linewidth}
		\centering
		\includegraphics[width=1.0\linewidth]{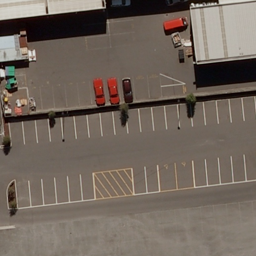}
	\end{minipage}
        \vspace{2pt}
	\begin{minipage}{0.11\linewidth}
		\centering
		\includegraphics[width=1.0\linewidth]{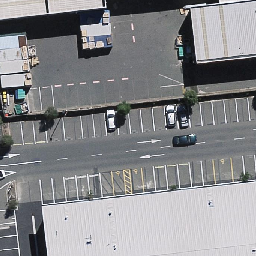}
	\end{minipage}
	\begin{minipage}{0.11\linewidth}
		\centering
		\includegraphics[width=1.0\linewidth]{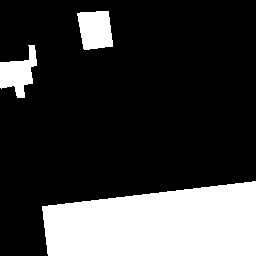}
	\end{minipage}
	\begin{minipage}{0.11\linewidth}
		\centering
		\includegraphics[width=1.0\linewidth]{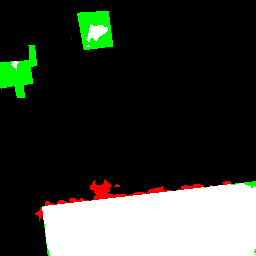}
	\end{minipage}
	\begin{minipage}{0.11\linewidth}
		\centering
		\includegraphics[width=1.0\linewidth]{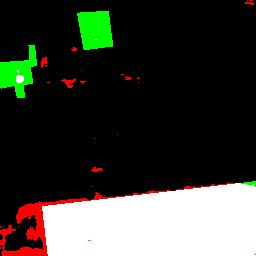}
	\end{minipage}
	\begin{minipage}{0.11\linewidth}
		\centering
		\includegraphics[width=1.0\linewidth]{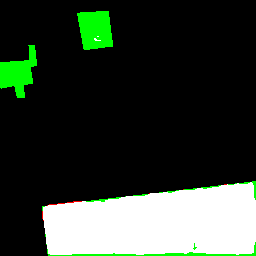}
	\end{minipage}
	\begin{minipage}{0.11\linewidth}
		\centering
		\includegraphics[width=1.0\linewidth]{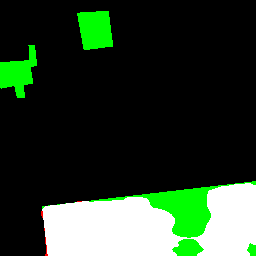}
	\end{minipage}
	\begin{minipage}{0.11\linewidth}
		\centering
		\includegraphics[width=1.0\linewidth]{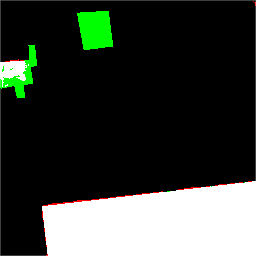}
	\end{minipage}
 	\begin{minipage}{0.11\linewidth}
		\centering
		\includegraphics[width=1.0\linewidth]{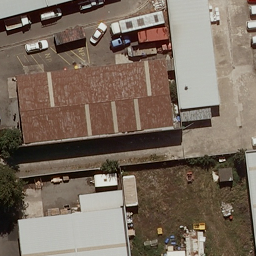}
	\end{minipage}
 \vspace{2pt}
	\begin{minipage}{0.11\linewidth}
		\centering
		\includegraphics[width=1.0\linewidth]{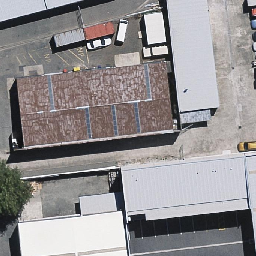}
	\end{minipage}
	\begin{minipage}{0.11\linewidth}
		\centering
		\includegraphics[width=1.0\linewidth]{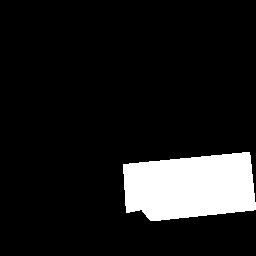}
	\end{minipage}
	\begin{minipage}{0.11\linewidth}
		\centering
		\includegraphics[width=1.0\linewidth]{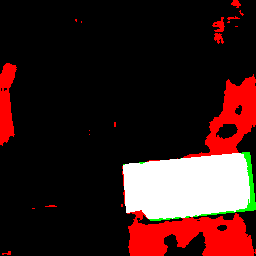}
	\end{minipage}
	\begin{minipage}{0.11\linewidth}
		\centering
		\includegraphics[width=1.0\linewidth]{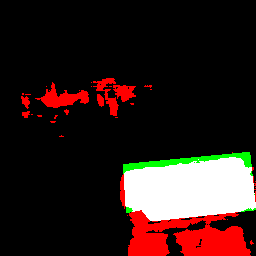}
	\end{minipage}
	\begin{minipage}{0.11\linewidth}
		\centering
		\includegraphics[width=1.0\linewidth]{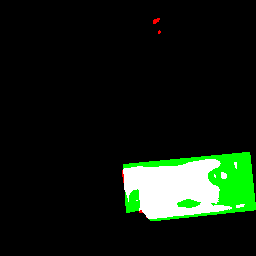}
	\end{minipage}
	\begin{minipage}{0.11\linewidth}
		\centering
		\includegraphics[width=1.0\linewidth]{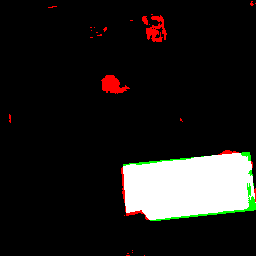}
	\end{minipage}
	\begin{minipage}{0.11\linewidth}
		\centering
		\includegraphics[width=1.0\linewidth]{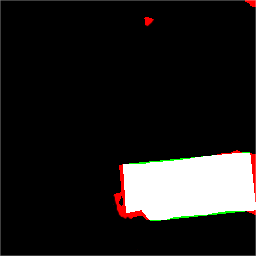}
	\end{minipage}
 \vspace{2pt}
 	\begin{minipage}{0.11\linewidth}
		\centering
		\includegraphics[width=1.0\linewidth]{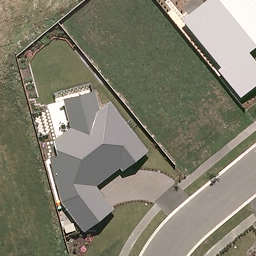}
	\end{minipage}
	\begin{minipage}{0.11\linewidth}
		\centering
		\includegraphics[width=1.0\linewidth]{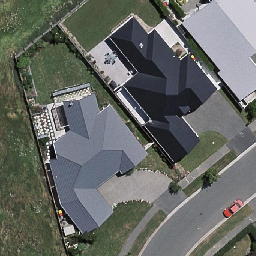}
	\end{minipage}
	\begin{minipage}{0.11\linewidth}
		\centering
		\includegraphics[width=1.0\linewidth]{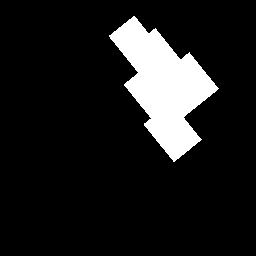}
	\end{minipage}
	\begin{minipage}{0.11\linewidth}
		\centering
		\includegraphics[width=1.0\linewidth]{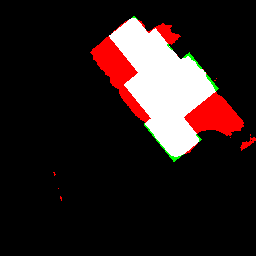}
	\end{minipage}
	\begin{minipage}{0.11\linewidth}
		\centering
		\includegraphics[width=1.0\linewidth]{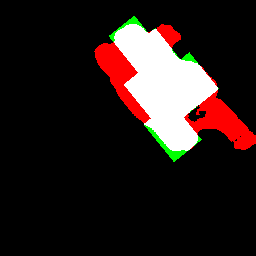}
	\end{minipage}
	\begin{minipage}{0.11\linewidth}
		\centering
		\includegraphics[width=1.0\linewidth]{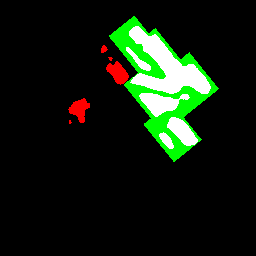}
	\end{minipage}
	\begin{minipage}{0.11\linewidth}
		\centering
		\includegraphics[width=1.0\linewidth]{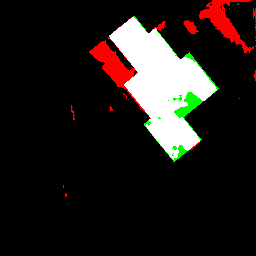}
	\end{minipage}
	\begin{minipage}{0.11\linewidth}
		\centering
		\includegraphics[width=1.0\linewidth]{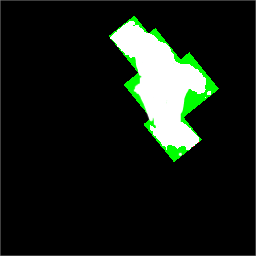}
	\end{minipage}
  	\begin{minipage}{0.11\linewidth}
		\centering
        \subfloat[]{\includegraphics[width=1.0\linewidth]{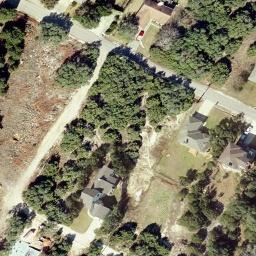}}
	\end{minipage}
	\begin{minipage}{0.11\linewidth}
		\centering
		\subfloat[]{\includegraphics[width=1.0\linewidth]{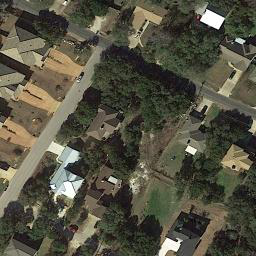}}
	\end{minipage}
	\begin{minipage}{0.11\linewidth}
		\centering
		\subfloat[]{\includegraphics[width=1.0\linewidth]{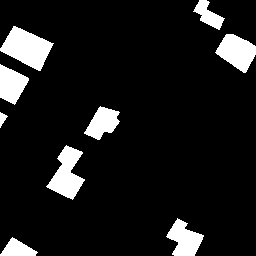}}
	\end{minipage}
	\begin{minipage}{0.11\linewidth}
		\centering
		\subfloat[]{\includegraphics[width=1.0\linewidth]{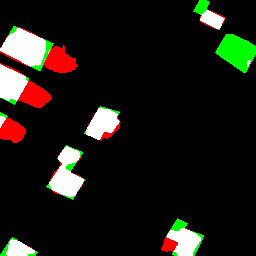}}
	\end{minipage}
	\begin{minipage}{0.11\linewidth}
		\centering
		\subfloat[]{\includegraphics[width=1.0\linewidth]{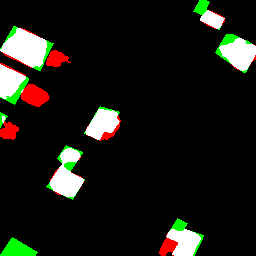}}
	\end{minipage}
	\begin{minipage}{0.11\linewidth}
		\centering
		\subfloat[]{\includegraphics[width=1.0\linewidth]{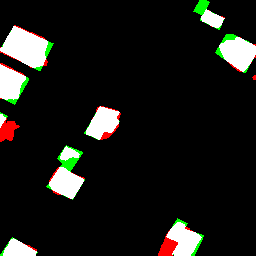}}
	\end{minipage}
	\begin{minipage}{0.11\linewidth}
		\centering
	\subfloat[]{\includegraphics[width=1.0\linewidth]{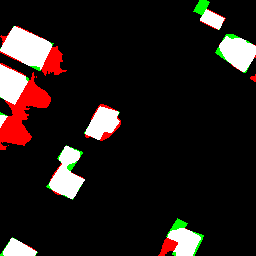}}
	\end{minipage}
	\begin{minipage}{0.11\linewidth}
		\centering
	\subfloat[]{\includegraphics[width=1.0\linewidth]{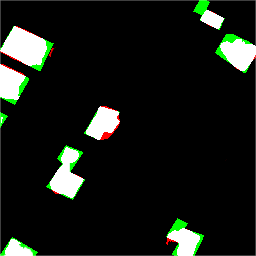}}
	\end{minipage}
        \caption{Comparison of our DDLNet (h) and representative RSCD methods: (d) FC-Siam-Di, (e) LGPNet, (f) SNUNet and (g) USSFC-Net, regarding visualized RSCD results, along with (a) $T_1$, (b) $T_2$ and (c) ground truth, on the WHU and LEVIR-CD test sets. Pixels are colored differently for better visualization (i.e., white for true positive, black for true negative, red for false positive, and green for false negative).}
	\label{Fig.2}
\end{figure*}

\subsection{Comparison and Analysis}
We conduct a comprehensive evaluation of our DDLNet on three benchmark datasets: WHU-CD, LEVIR-CD, and CLCD. Table \ref{tab:1} presents the performance comparison with current state-of-the-art methods, revealing that DDLNet achieves F1 scores of 90.2\% (WHU-CD), 90.6\% (LEVIR-CD), and 72.44\% (CLCD), showcasing significant improvements across all datasets. This performance superiority can be attributed to the effective feature optimization by our FEM and SRM modules, utilization of multiple frequency components for enhanced changes of interest, improved feature representation in pixel space, and fine-grained information capture through feature fusion, facilitating the recovery of spatial details in change representations.

Furthermore, we conduct ablation experiments on WHU-CD datasets. Table \ref{tab:2} presents that the base model without FEM and SRM modules achieve F1 scores of 87.96\% on WHU-CD dataset. The introduction of the FEM or SRM module significantly improve performance, with our complete DDLNet achieving the highest F1 of 90.2\%, confirming the effectiveness of these modules in enhancing change detection performance.

Additionally, as shown in Table \ref{tab:3}, we validate the results using $n$ frequency components, where we set $n$ to be 4, 8, 16, and 32. It can be observed that the configuration with 16 frequency components achieve optimal performance. Furthermore, as depicted in Table \ref{tab:4}, DDLNet, with 12.67M parameters and 7.35G Flops, demonstrate superior efficiency, requiring fewer parameters and boasting the lowest computational cost compared to other methods.

We visually validate the change detection results on three constructed datasets in Fig.\ref{Fig.2}. The comparison plots illustrate that DDLNet outperforms other models by effectively leveraging the FEM and SRM modules to extract high-quality change information and enhance change object boundaries.

\section*{Conclusion}

In this paper, we propose a DDLNet for change detection in remote sensing images. The proposed DDLNet is based on dual-domain learning in the frequency and spatial domains. Our DDLNet captures effective frequency components in features using the discrete cosine transform, thereby enhancing the changes of interest between dual temporal images. Additionally, through the spatial-domain recovery module, it fuses spatiotemporal features to restore spatial details of the change representation. We evaluate DDLNet on three public change detection datasets, and the method achieves state-of-the-art performance on all benchmark datasets, demonstrating its effectiveness in remote sensing change detection.

% \vspace{12pt}
% \color{red}
% IEEE conference templates contain guidance text for composing and formatting conference papers. Please ensure that all template text is removed from your conference paper prior to submission to the conference. Failure to remove the template text from your paper may result in your paper not being published.

\end{document}